\documentclass[runningheads, anonymous]{llncs}
\usepackage{graphicx}
\usepackage{hyperref}
\usepackage{booktabs}
\usepackage{tabularx}
\usepackage{makecell}
\usepackage{amsmath}
\usepackage{amssymb}
\usepackage{multirow}
\usepackage{marvosym}

\hypersetup{
    colorlinks=true,      % 将链接文字设置为彩色，而不是默认的方框
    linkcolor=blue,       % 内部交叉引用链接颜色（如图表引用）
    citecolor=blue,       % 文献引用链接颜色
    urlcolor=blue,        % URL链接颜色
    allcolors=blue        % 统一所有链接为蓝色（可选）
}

\begin{document}
\title{PressureMesh: 3D Human Mesh Estimation from Multi-Device
Pressure Images}
%
%\titlerunning{Abbreviated paper title}
% If the paper title is too long for the running head, you can set
% an abbreviated paper title here
%
% \author{ANONYMOUS AUTHOR(S)}

\author{Changhai Ma \and Ziyu Wu \and Yunkang Zhang \and Fangting Xie \and Mengting Niu \and Heyu Ding \and Quan Wan \and Jiayue Yuan \and Boyan Liu \and Yi Ke \and Xiaohui Cai\textsuperscript{\Letter}}

\institute{University of Science and Technology of China, Hefei, China\\
\email{mach1126@mail.ustc.edu.cn\\ caixiaohui@ustc.edu.cn}}

% %
% \authorrunning{F. Author et al.}
% % First names are abbreviated in the running head.
% % If there are more than two authors, 'et al.' is used.
% %
% \institute{Princeton University, Princeton NJ 08544, USA \and
% Springer Heidelberg, Tiergartenstr. 17, 69121 Heidelberg, Germany
% \email{lncs@springer.com}\\
% \url{http://www.springer.com/gp/computer-science/lncs} \and
% ABC Institute, Rupert-Karls-University Heidelberg, Heidelberg, Germany\\
% \email{\{abc,lncs\}@uni-heidelberg.de}
%
\maketitle              % typeset the header of the contribution
\begin{abstract}
Human pose monitoring is crucial in fields such as rehabilitation assessment and human-computer interaction. Due to its privacy-preserving nature, pressure-based human pose monitoring has become a primary approach for unobtrusive sensing. However, existing methods are generally limited to a single device, which restricts the effective monitoring range. To address this limitation, we propose MDP-Net, an end-to-end network capable of directly estimating human meshes from temporal pressure data across multiple devices. We introduce a multimodal fusion mechanism inspired by the Mixture of Experts (MoE) framework to achieve effective complementarity and enhancement of cross-device pressure information. To support the training and evaluation of MDP-Net, we constructed MDP, a high-quality multi-device temporal pressure dataset that includes various pose labels such as 2D/3D joints and human meshes. Experimental results demonstrate that MDP-Net achieves a joint position error of 12.6 cm on the MDP dataset. These results prove that fusing multi-device pressure information is an effective and promising new solution for daily human pose monitoring.

\end{abstract}
\section{Introduction}

Accurate human pose sensing is essential for health monitoring, rehabilitation, and human-computer interaction. Prevailing RGB-based methods rely on visual cues (e.g., 2D keypoints~\cite{smplify}, contours~\cite{pymaf}) and are sensitive to lighting, occlusion, and privacy concerns. Alternatives such as radar~\cite{radar1,radar2}, Wi-Fi~\cite{wifi1,wifi2} sensing raise privacy issues due to penetration capability, while inertial measurement units (IMUs) suffer from limited battery life and wearing discomfort.

In daily activities, the human body frequently contacts surfaces such as carpets, chairs, and beds, generating pressure signals that encode rich pose information. Pressure-based sensing offers advantages including lighting invariance, penetration of soft covers, and natural privacy preservation. Although prior studies~\cite{pimesh,pihmr,pidhmr,Intelligentcarpet,motionpro,Cavatar,zmj} have used pressure sensor arrays in single devices like mattresses or carpets, they are limited in sensing range and fail to capture full-body poses in multi-surface interactions (e.g., sitting on a chair with feet on a carpet). Therefore, fusing pressure data from multiple devices presents a promising direction toward more accurate and robust pose estimation.

% To explore the feasibility of multi-device pressure information fusion, we propose MDP-Net: a  human mesh reconstruction method based on multi-device temporal pressure signals. This method co-processes asynchronous pressure data from various contact surfaces—including beds, carpets, and chairs—to achieve end-to-end joint estimation of 3D human pose and shape. To address the challenges posed by diverse pressure distributions across devices and the difficulty of achieving effective information complementarity, we designed a two-stage core mechanism. At the input stage, we employ a device-specific normalization strategy to unify the scales and distributions of data from different sources. In the feature fusion stage, inspired by the Mixture of Experts (MoE)~\cite{moe} framework, we designed a dynamically learnable multi-source fusion module. This module utilizes a lightweight gating network to automatically parse input data patterns, dynamically activating and weighting the most relevant expert branches. This enables adaptive selection and optimal fusion of cross-device information.
% To address data heterogeneity, we first apply device-specific data preprocessing methods to align distributions and suppress noise (see Supplement).
To explore the feasibility of multi-device pressure information fusion, we propose MDP‑Net, an end‑to‑end network that processes temporal pressure data from bed, carpet and chair, then jointly estimates pose and shape and reconstructs 3D human mesh. The architecture is designed to address two core challenges: heterogeneous data fusion and pose recovery from information‑sparse pressure data. To handle data heterogeneity,  we first apply device-specific data preprocessing methods to align distributions and suppress noise (see Sup. Mat.). We then introduce a Mixture‑of‑Experts (MoE)‑based fusion module that dynamically selects and enhances complementary cross‑device features. To address information sparsity, a dedicated pose regressor decodes the fused features into pose parameters. This modular design separates responsibilities: the fusion module integrates multi‑device information, while the regressor learns the mapping from pressure features to pose, improving accuracy in low‑information‑density settings.

To support the training and validation of MDP‑Net, we introduce MDP, the first large‑scale Multi‑device Pressure dataset, capturing synchronized pressure and pose across multiple surfaces. It covers three common contact devices—bed, carpet, and chair—in daily scenarios. Data were collected from 12 subjects, including 672K multi‑view RGB frames and 96K synchronized multi‑device pressure sequences (288K pressure images in total). Each sample is annotated with accurate 2D keypoints, 3D keypoints and parametric human meshes (SMPL~\cite{smpl}), offering a comprehensive benchmark for multi‑device pressure‑based pose estimation.

In summary, our contributions are threefold:
\begin{itemize}
\item[1] We propose Multi-Device Pressure Network (MDP-Net), the first approach to jointly estimate human shape from multi-device sequential pressure data. Our method can directly predict 3D human meshes from pressure image sequences collected by beds, chairs, and carpets.
% We propose MDP‑Net, the first method to estimate 3D human pose and shape directly from multi‑device temporal pressure data. 

\item[2] We collect the first multi‑device pressure dataset, containing 672K multi-view RGB frames and 96K synchronized pressure samples (288K pressure images) from 12 subjects, with 2D/3D keypoints and 3D mesh (SMPL) annotations.
% We construct a new multi-device pressure dataset, which includes data from 12 participants. The dataset contains a total of 672K multi-view RGB images and 96K paired samples comprising 288K multi-device pressure maps, along with diverse ground-truth human pose representations, including 2D keypoints, 3D keypoints, and 3D meshes.

% We collect the first multi‑device pressure dataset, containing 672K multi‑view RGB frames and 96K synchronized pressure samples (288K pressure maps) from 12 subjects, with 2D/3D keypoints and 3D mesh (SMPL) annotations.

\item[3] MDP‑Net achieves 12.6cm MPJPE and 14.5cm MPVE on MDP dataset, outperforming single‑device baselines across standing, sitting, and lying subsets. Furthermore, we conducted extensive ablation studies to explore the optimal network configuration.

% MDP-Net achieves an average joint position error of 12.6 cm  and an average vertex error of 14.5 cm on our dataset. Our method consistently outperforms single-device approaches on the standing, sitting, and lying pose subsets. Furthermore, we conducted extensive ablation studies to explore the optimal network configuration. 

\end{itemize}

\section{Related Work}

\subsection{Human mesh estimation from pressure images}
% Significant contributions have been made by several research teams in the field of pressure-based human pose estimation. Pressure sensors embedded in beds have been successfully applied to pose estimation tasks~\cite{pimesh,pihmr,pidhmr}. Among these, PiMesh~\cite{pimesh} proposed an end-to-end solution that directly utilizes temporal pressure maps as input to estimate SMPL model parameters. Building upon traditional end-to-end approaches, PidHMR~\cite{pidhmr} innovatively introduced diffusion models as prior knowledge to assist in shape estimation, thereby enhancing both accuracy and stability. In contrast, research on chair-based pose estimation remains relatively sparse. To the best of our knowledge, Zhao et al.~\cite{zmj} was among the first to estimate 3D human poses directly from chair pressure data; however, this method was limited to 3D skeletal structures and failed to reconstruct a full SMPL model. Regarding carpet-based sensing, Luo et al.~\cite{Intelligentcarpet} developed an innovative approach using pressure-sensitive carpets; while achieving promising results, their work was similarly restricted to 3D skeletal models. Cavatar~\cite{Cavatar} subsequently improved upon Luo’s work, successfully achieving full SMPL mesh estimation from carpet data.

Research on pressure‑based pose estimation has advanced notably across different surfaces. For bed‑embedded sensors, end‑to‑end methods such as PiMesh~\cite{pimesh,pihmr} directly regress SMPL parameters from pressure images, while PidHMR~\cite{pidhmr} further employs diffusion priors to improve accuracy. In chair‑based sensing, Zhao et al.~\cite{zmj} pioneer 3D pose estimation from chair pressure, yet only produce skeletal outputs. For carpets, Luo et al.~\cite{Intelligentcarpet} introduce a pressure‑sensitive carpet for 3D pose estimation, later extended by Cavatar~\cite{Cavatar} to full SMPL mesh recovery.

% Despite significant progress in pressure-based human pose estimation, existing methods are generally constrained by a single-device sensing paradigm. This paradigm is ill-suited for complex daily scenarios where the human body interacts simultaneously with multiple surfaces, such as beds, chairs, and carpets. Due to the limited sensing range, such approaches often fail to capture comprehensive pressure distribution information, thereby restricting their application potential in intricate real-life settings. In contrast, our method leverages multi-device pressure information to not only expand the physical sensing range but, more importantly, to exploit the information complementarity between different devices, ultimately achieving more accurate and robust human pose estimation.

Despite these advances, all existing methods rely on a single‑device setup, which limits their applicability in daily multi‑surface interactions (e.g., sitting on a chair with feet on a carpet). The restricted sensing range of a single device often leads to incomplete pressure coverage, hindering full‑body pose estimation. In contrast, our work leverages multi‑device pressure fusion, expanding the sensing range and, more importantly, exploiting inter‑device complementarity to achieve more accurate and robust pose estimation.

\subsection{Public pressure dataset}
Corresponding pressure-pose datasets have been developed for different support surfaces. For bed-based monitoring, Pmat~\cite{pmat} collects approximately 18K pressure frames from 13 subjects, while SLP~\cite{slp} scales to 15K samples from 109 subjects under different covering conditions. TIP~\cite{pimesh} provides 152K temporal pressure frames with high-quality SMPL parameters generated via Cliff~\cite{cliff} and optimization. For chair-based monitoring, Zhao et al.~\cite{zmj} collect a temporal chair pressure dataset with 3D keypoint annotations. In carpet-based monitoring, Luo et al.~\cite{Intelligentcarpet} capture carpet pressure data with 3D keypoints. Cavatar~\cite{Cavatar} extends it with pseudo-3D mesh labels, and MotionPro~\cite{motionpro} records 12.4M frames synchronized with a MoCap system. A summary of these datasets is given in Table~\ref{tab1}.

% While the aforementioned datasets have significantly propelled research in pressure-based sensing within their respective domains, they are invariably confined to single-device acquisition. To the best of our knowledge, there is currently no publicly available dataset that provides synchronized pressure information from multiple devices. Consequently, to address this gap, we developed a synchronized acquisition system for multi-device pressure and human poses, and introduced the first multi-device pressure dataset (MDP).

Despite their contributions, all existing datasets are confined to single‑device acquisition. To our knowledge, no public dataset provides synchronized pressure signals from multiple devices. To bridge this gap, we develop a synchronized multi‑device pressure‑pose acquisition system and present the first multi‑device pressure dataset (MDP).

\vspace{-10pt}
\begin{table}
\caption{List of Current Public Pressure Datasets. SV: Single-View, DV: Dual-View, MV: Multi-View, B: Bed, Ch: Chair, Ca: Carpet, Kp: Keypoints}\label{tab1}
\centering
\begin{tabular}{ccccccc}
\Xhline{2\arrayrulewidth}
Dataset & Vision & Device  & Human Body & Subject  & Frame & Temporal  \\
\Xhline{2\arrayrulewidth}
PMat~\cite{pmat} & - & B & - & 13 & 18K & Yes\\
SLP~\cite{slp} & SV RGBD & B & 2D Kp & 109 & 14.7K & No\\
TIP~\cite{pimesh} & SV RGBD & B & SMPL & 9 & 152K &  Yes\\
Zhao et al.~\cite{zmj} & MV RGB & Ch & 3D Kp & 14 & 180K &  Yes \\
Luo et al.~\cite{Intelligentcarpet} & DV RGB & Ca & 3D Kp & 10 & 180K &  Yes \\
MotionPro~\cite{motionpro} & MV RGB & Ca & SMPL & 70 & 12.4M & Yes  \\
\hline
Ours & MV RGB & B, Ch, Ca & SMPL & 12 & 288K &  Yes \\ 
\Xhline{2\arrayrulewidth}
\end{tabular}
\end{table}
\vspace{-20pt}

\section{Dataset}

% The construction of large-scale, high-quality annotated datasets is fundamental for the effective training and reliable evaluation of deep learning models. This section introduces a novel multimodal dataset developed for this purpose, which provides synchronized multi-view RGB video sequences and multi-device temporal pressure data. The dataset is annotated with high-precision pseudo-labels, including 2D/3D human keypoints and 3D human meshes. These 3D human mesh labels are derived from multi-view video reconstruction and will serve as the foundation for subsequent supervised learning.image

Large‑scale, high‑quality annotated datasets are essential for training and evaluating deep learning models. To this end, we construct a multimodal dataset that provides synchronized multi‑view RGB images and multi‑device temporal pressure images. Each sample is annotated with high‑precision pseudo labels, including 2D/3D keypoints and 3D human meshes reconstructed from multi‑view images, which serve as the supervision signal for subsequent learning.

\subsection{Hardware Configuration and Environment Setup.}
\begin{figure}[htbp]
\centerline{\includegraphics[width=1.0\linewidth]{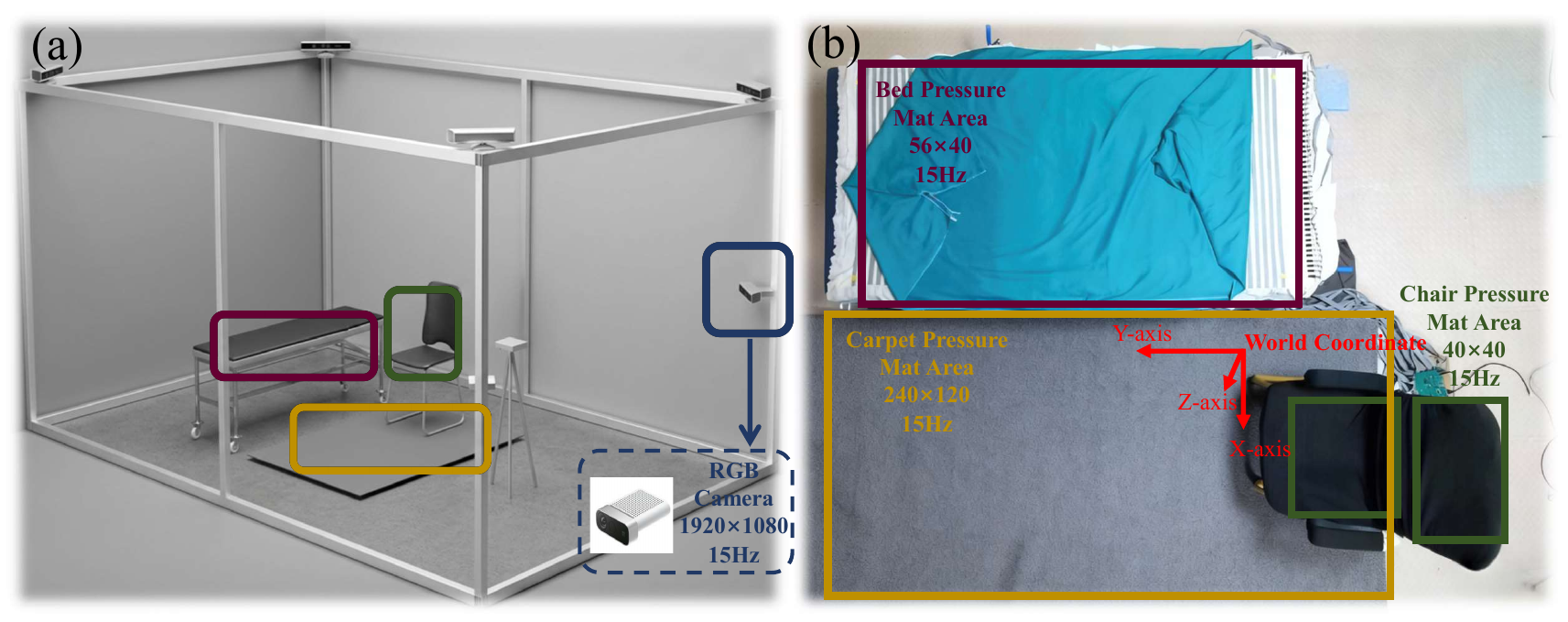}}
\caption{The architecture of our system for dataset collection. (a) shows the overall architecture of our system, comprising seven cameras and three pressure-sensing devices. (b) displays the specific coverage areas and sensor parameters of the pressure-sensing devices. Rectangular boxes of the same color in (a) and (b) represent the same device.}
\label{env}
\end{figure}

 01
To construct a multimodal data acquisition platform, we set up an experimental area measuring $6m \times 3m \times 2.5m$. Seven Azure Kinect DK RGB-D cameras were evenly distributed around the perimeter of the area to synchronously capture multi-view high-definition RGB images. A diagram of the actual experimental setup is shown in Fig~\ref{env}. Common furniture items—namely a single bed, a chair, and a carpet—were placed in the center of the area. Pressure sensing devices were installed beneath each of these items; their specific parameters are listed below:
\begin{itemize}
\item[$\bullet$]\textbf{Bed Pressure Mat:} The active sensing area is $1.96m \times 0.96m$, composed of a $56 \times 40$ grid of sensing elements with a spatial resolution of $3.11cm \times 1.95cm$.
\item[$\bullet$]\textbf{Seat Pressure Mat:} Comprising two sensor arrays, each with a configuration of $19 \times 34$ elements. The total dimensions are $0.37m \times 0.33m$, with an element pitch of $1.96 cm \times 0.96cm$.
\item[$\bullet$]\textbf{Carpet Pressure Mat:} Features a sensing area of $1.2m \times 2.4m$, integrating a dense grid of $238 \times 120$ sensing elements with a fine pitch of $1cm \times 1cm$.
\end{itemize}

\vspace{-15pt}  % 减少图片上方的间距，数值可调整
\vspace{-20pt}  % 减少图片下方的间距，数值可调整

\subsection{Label Generation}

\subsubsection{Preliminary: SMPL Model.} SMPL~\cite{smpl} is a widely-used parametric human model defined as a function $M=(\theta, \beta) \in \mathbb{R}^{6890\times3}$, which outputs a 3D mesh from pose parameters $\theta\in\mathbb{R}^{72}$ (axis-angle rotations of 24 body joints) and shape parameters $\beta\in\mathbb{R}^{10}$ (encoding height, weight, and other body shape variations).

% SMPL~\cite{smpl} is the most widely used parametric human body model. It provides a function, $M=(\theta, \beta)$, which takes pose parameters $\theta$ and shape parameters $\beta$ as input and returns a 3D human mesh $M\in \mathbb{R}^{6890\times3}$. The pose parameters $\theta\in\mathbb{R}^{72}$ represent the relative rotations of the 24 body joints in an axis-angle representation, resulting in a 72-dimensional vector. The shape parameters $\beta\in\mathbb{R}^{10}$ are associated with factors such as the subject's height and weight, controlling the body shape of the model.

\subsubsection{2D Keypoints Label Generation.}
% We utilize OpenPose to extract 2D joints from multiple viewpoints. However, due to factors such as occlusion and illumination, 2D joint detection in certain frames may be biased or missing. To address this, we designed a two-stage refinement pipeline to enhance label quality. In the first stage, we leverage the complementarity of multiple views: while the data quality of one viewpoint may degrade due to environmental factors, others may remain unaffected. We re-project the generated 3D joints back onto each perspective to perform projection consistency correction on the 2D joints. The second stage involves manual verification, where joints that still exhibit significant deviations are manually corrected based on temporal motion continuity and natural human movement priors. Typical examples of these corrections are shown in Fig~\ref{img:correct}.

\begin{figure}[htbp]
\centerline{\includegraphics[width=1.0\linewidth]{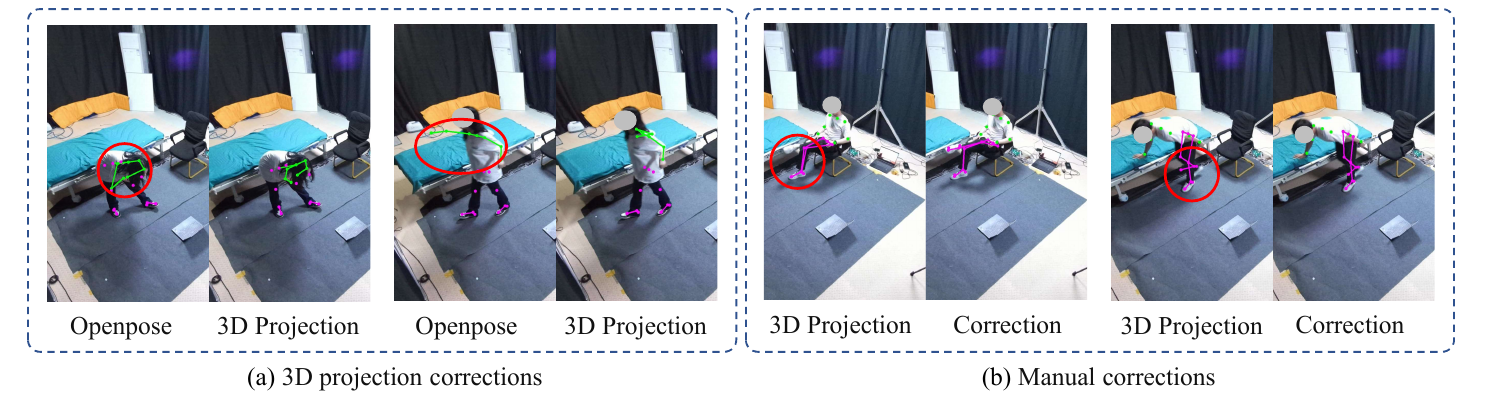}}
\caption{Some examples of 3D projection and manual correction. We highlight the correction positions by red ellipses.}
\label{img:correct}
\end{figure}

We extract 2D keypoints from multiple views using OpenPose. To address detection errors or missing keypoints caused by occlusion or illumination, we employ a two-stage refinement pipeline. First, we leverage multi-view complementarity: 3D keypoints are triangulated and reprojected to each view to correct inconsistencies. Second, remaining keypoints with large deviations are manually adjusted based on temporal motion continuity and natural human movement priors. Some example corrections are shown in Fig.~\ref{img:correct}.

\vspace{-15pt}
\vspace{-20pt}

\subsubsection{3D Keypoints and SMPL Label Generation.}
Initially, we performed camera calibration using MATLAB to unify the world coordinate origins and obtain the intrinsic and extrinsic parameters for all cameras. Subsequently, based on these calibration results, the 3D spatial positions of the human joints were reconstructed via triangulation.

Following the approach of EasyMocap~\cite{easymocap}, we adopt a multi-view optimization-based SMPLify~\cite{smplify} approach, utilizing the aforementioned 2D and 3D joints as supervision to generate SMPL models. This optimization-based method updates the SMPL parameters through gradient descent to minimize the objective function $L$. The objective function $L$ is defined as follows:
\begin{align}
L_{total}(\theta, \beta, t)=\lambda_{2d}\sum \limits_{n=1}^{7}L_{2d}+\lambda_{3d}L_{3d}+\lambda_{smooth}L_{smooth}+\lambda_{reg}L_{reg}
\end{align}
Where $L_{2d}$ and $L_{3d}$ denote the $L_2$ distance losses between the predicted 2D and 3D joints and their corresponding ground truth, respectively. $L_{reg}$ is a regularization constraint imposed on the human pose parameters. $L_{smooth}$  is the temporal smoothing loss, which is designed to constrain the variations in human shape and pose between consecutive frames. $L_{smooth}$ consists of two parts: $L_{pose}$ and $L_{shape}$.
\begin{equation}
\begin{aligned}
    L_{pose}=\sum\limits_{i=1}^{N-1}\|J_{3d}(i)-1/2(J_{3d}(i-1)+J_{3d}(i+1)\|^2 _2+ \\ \sum\limits_{i=1}^{N-2}\|\theta(i)-1/3(\theta(i-1)+\theta(i)+\theta(i+1)\|^2_2
\end{aligned}
\end{equation}
% \vspace{-7pt}
\begin{align}
    L_{shape} = \sum\limits_{i=1}^N\|\beta(i+1)-\beta(i)\|^2
\end{align}

\begin{figure}[htbp]
\centerline{\includegraphics[width=1.0\linewidth]{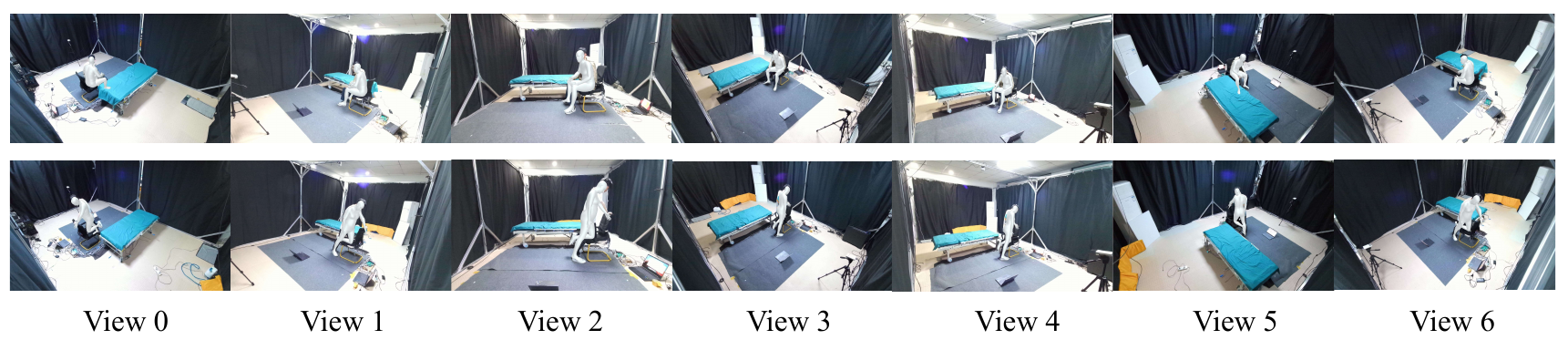}}
\caption{Some examples of the generated SMPL labels.}
\label{img:smplshow}
\end{figure}

For the shape constraint, since the input sequence originates from the same subject, the shape parameters $\beta$ are expected to remain constant across all frames. As for the pose smoothing constraint, the smoothing terms for both 3D joints $J_{3d}$ and pose parameters $\theta$ are minimized to suppress jitter. Partial results are illustrated in Fig~\ref{img:smplshow}.

\vspace{-15pt}
\vspace{-20pt}

\section{Method}
% Temporal pressure-based pose estimation preserves privacy but presents inherent challenges: pressure images are indirect, sparse 2D projections with lower information density than RGB images, making pose decoding difficult. Scaling from single to multiple devices shifts the core challenge from interpreting to fusing data—i.e., harmonizing multi-source signals with differing resolutions and noise for effective spatiotemporal alignment and complementarity. To overcome this, we introduce a Mixture of Experts (MoE)-inspired model, MDP-Net, an end-to-end network for multi-device temporal pressure data that estimates 3D human meshes directly from distributed pressure inputs. The overall architecture is shown in Fig~\ref{img:pipeline}, details are provided in subsequent sections.

Temporal pressure-based pose estimation offers privacy advantages but faces two inherent challenges: pressure images are indirect, sparse 2D projections with lower information density than RGB, making pose recovery difficult; furthermore, extending from single-device to multi-device sensing shifts the core problem from interpreting data to fusing it—i.e., aligning multi-source signals with varying resolutions and noise for spatiotemporal complementarity. To address these, we introduce MDP-Net, an end-to-end network based on a Mixture of Experts (MoE) design that directly estimates 3D human meshes from distributed pressure inputs. The overall architecture is illustrated in Fig.~\ref{img:pipeline}, details are provided in subsequent sections.

\subsection{Network Structure}
% Fig~\ref{img:pipeline} illustrates the overall pipeline of the proposed MDP-Net framework. The method takes temporal multi-device pressure data as input and achieves end-to-end 3D human mesh prediction. We process pressure data from different devices through separate pathways. Specifically, the input pressure sequences are first fed into a ResNet-18~\cite{resnet} backbone for feature extraction, followed by a three-layer Transformer~\cite{transformer} encoder (enhanced with positional encoding for temporal awareness) to capture inter-frame dependencies. The features from each pathway are then concatenated and passed through a routing layer that adaptively activates relevant expert branches within the Mixture of Experts (MoE) architecture. Finally, the outputs of the expert branches are aggregated and fed into a multi-layer perceptron-based regressor to generate the final 3D human mesh.

\begin{figure}[htbp]
\centerline{\includegraphics[width=1.0\linewidth]{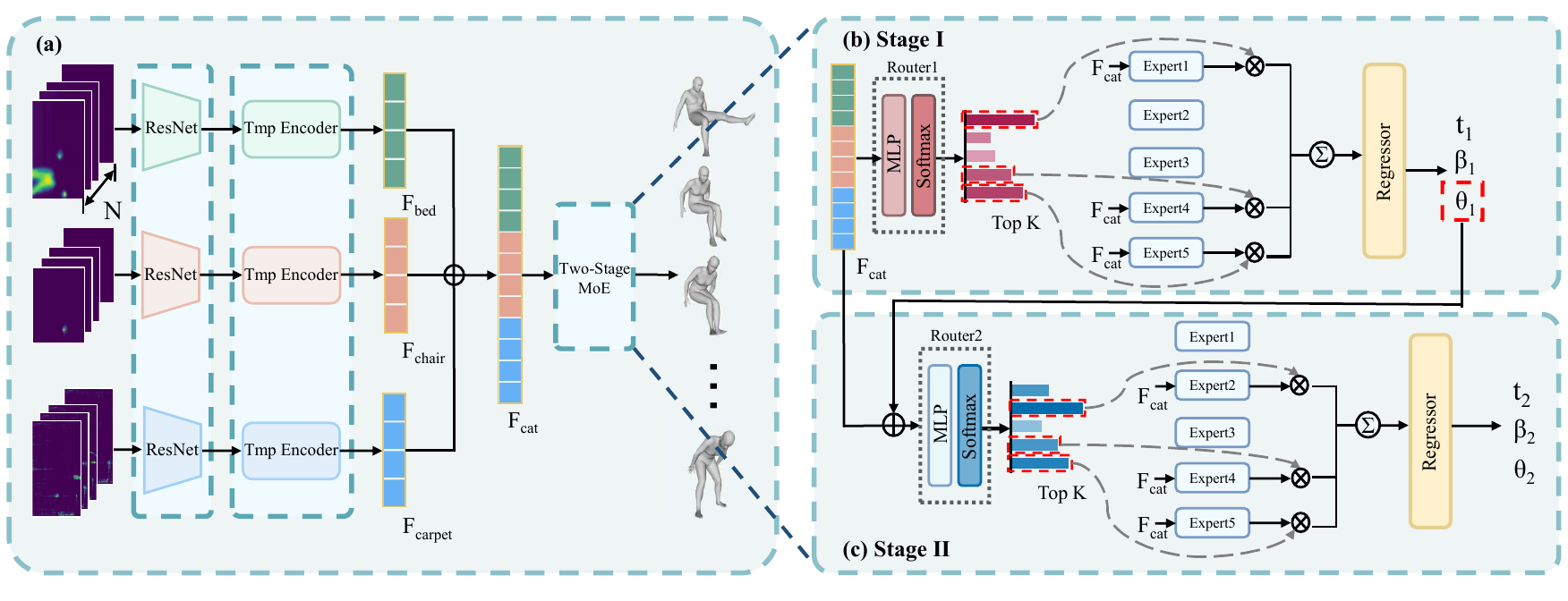}}
\caption{The overview framework of our proposed network MDP-Net. (a) shows MDP-Net's overall architecture, processing multi-device temporal pressure inputs to generate SMPL model predictions. (b) and (c) represent the two-stage MoE design, where the expert networks parameters are shared between the two stages.}
\label{img:pipeline}
\end{figure}

Fig.~\ref{img:pipeline} depicts the overall pipeline of MDP‑Net. The framework takes temporal multi‑device pressure data as input and performs end‑to‑end 3D human mesh regression. Pressure sequences from each device are processed in separate pathways: a ResNet‑18~\cite{resnet} backbone extracts spatial features, followed by a 3‑layer Transformer~\cite{transformer} encoder (with positional encoding) to model temporal dependencies. The resulting features are concatenated and routed through a Mixture of Experts (MoE) layer, which dynamically activates relevant experts. the outputs of the expert branches are aggregated and fed into a multi-layer perceptron-based regressor to generate the final 3D human mesh.

% \vspace{-15pt}
% \vspace{-20pt}

% \subsection{Feature Extractor and Temporal Encoder}
% We employ ResNet-18 as the backbone network to extract spatial features from the original pressure images. To accommodate the single-channel pressure input data, we modify its first convolutional layer to accept 1 channel instead of the standard 3 channels (for RGB), while keeping the rest of the architecture unchanged. For temporal modeling, we adopt a three-layer Transformer encoder to capture the dynamic dependencies within the pressure sequence of each device. This encoder takes the frame-by-frame feature sequences extracted by ResNet as input, models long-range temporal contexts through its self-attention mechanism, and outputs temporally-aware enhanced feature representations for further processing by the subsequent multi-device fusion module.

\subsection{Feature Fusion Module}
To effectively integrate pressure information from multiple devices and leverage their spatiotemporal complementarity, we draw inspiration from the divide-and-conquer philosophy of the Mixture of Experts (MoE) model and propose an adaptive selective fusion mechanism.

The core of this module is a gating network, composed of an MLP and a softmax layer. It takes the initially concatenated features $F_{cat}$ input and produces a set of expert weights $G=[g_1,g_2,...,g_N]$. The computation process is shown in Equation~\ref{gate}:
\begin{align}
\label{gate}
    G=Gate_1(F_{cat})=softmax(W \cdot F_{cat} + b)
\end{align}
where $N$ is the total number of experts. These weights are not used to directly select experts, but rather serve as signals for dynamic feature selection and recombination, indicating how the feature transformation patterns provided by different experts should be weighted and combined for the current input pose.

Each expert $E_i$ is itself an MLP, and all experts take the same $F_{cat}$ input. Their design objective is to enable different experts to learn distinct decoding methods or to enhance fused features differently. Ultimately, the output features from all expert networks are aggregated via a weighted sum based on the gating weights $G$, thereby producing a refined and pose-adaptive enhanced fusion feature. This feature is then fed into a regressor $Regressor_1$, composed of MLP and Dropout layers, to predict the initial SMPL parameters $\theta_1$, $\beta_1$, and $t_1$. This process can be expressed as:
\begin{align}
    \theta_1,\beta_1,t_1 = Regressor_1(\sum\limits_{i=1}^{N} g_i \cdot E_i(F_{cat}))
\end{align}

While the initial estimation can roughly capture the overall pose, subtle pressure distributions (such as local pressure under the heels) may be weakened during the complex feature fusion process. To fully leverage these subtle yet critical cues, this study introduces a second MoE process to iteratively refine the results.

At this stage, the initial SMPL pose parameters $\theta_1$ and the original concatenated features $F_{cat}$ are concatenated and jointly used as input to the new-stage gating network. The rationale behind this design is that the initial human mesh estimate provides valuable geometric context (such as the approximate orientation of limbs and the likely locations of contact points), which can assist the gating network in making more precise decisions.

The second-stage MoE process follows the same weighted summation mechanism as the first stage, but its input incorporates more comprehensive contextual information. Finally, the refined SMPL parameters $\theta_2$, $\beta_2$, and $t_2$ are produced by the $Regressor_2$.
\begin{align}
    G=Gate_2(Contact(F_{cat}, \theta_1))
\end{align}
\vspace{-7pt}
\begin{align}
    \theta_2,\beta_2,t_2 = Regressor_2(\sum\limits_{i=1}^{N} g_i \cdot E_i(F_{cat})) 
\end{align}

\subsection{Regressor}
Inspired by the work of Cliff~\cite{cliff}, we design independent pose regressor within the network to decode SMPL parameters from the fused features. This design follows a modular philosophy of separation of responsibilities: the Mixture-of-Experts module focuses on learning interaction and correlation patterns across multiple devices to generate more discriminative fused features, while the pose regressor is dedicated to mapping these features to the human pose parameter space with physical constraints.

The pose regressor consists of three fully connected layers and Dropout layers. Its final output feature dimension is $(B \times L, 157)$, where $B$ is the batch size and $L$ is the sequence length. The first 144 dimensions of this feature correspond to the pose parameters $\theta$,  the next 10 dimensions correspond to the shape parameters $\beta$, and the last 3 dimensions correspond to the global translation $t$, together forming the complete SMPL parameter output.

\subsection{Loss Function}
The overall loss function of the proposed MDP-Net can be expressed as follows:
\begin{align}
L_{total}=\lambda_{3d}L_{3d}+\lambda_{smpl}L_{smpl}+\lambda_{v}L_{v}
\end{align}
where $L_{3d}$ minimizes L1 loss between estimated and ground truth 3D joints regressed from SMPL vertices, $L_{smpl}$ presents the deviations between the estimated SMPL parameters and ground truths, $L_v$ minimizes L1 loss between estimated and ground truth SMPL vertices. Each term is calculated as:
\begin{align}
    L_{3d} =\frac{1}{N} \sum\limits_{i=1}^N\sum\limits_{j=1}^J\|J_{3d}(i, j) - J^{GT}_{3d}(i, j)\|
\end{align}
\vspace{-7pt}
\begin{align}
    L_{v} = \frac{1}{N}\sum\limits_{i=1}^N\sum\limits_{j=1}^{6890}\|J_{v}(i, j) - J^{GT}_{v}(i, j)\|
\end{align}
\vspace{-7pt}
\begin{align}
    L_{smpl}=\frac{1}{N}\sum\limits_{i=1}^N(\omega_s\|\beta(i)-\hat{\beta}(i)\|+\omega_p\|\theta(i)-\hat{\theta}(i)\|+\omega_t\|t(i)-\hat{t}(i)\|)
\end{align}
where $\hat{x}$ represents the ground truth for the corresponding estimated variable $x$,$N$ is the input sequence length,
 and $\lambda_{3d}, \lambda_{smpl}, \lambda_{v}, \omega_s, \omega_p, \omega_t$ are the weights of individual loss function.

\section{Experiments}
\subsection{Dataset and Evaluation Metrics}
\subsubsection{Dataset.}
% To comprehensively evaluate the model's generalization capability in real-world scenarios, we systematically validated its performance using two distinct data splitting strategies: the unseen-group and unseen-subject settings.

% In the unseen-group setting, each subject performed a total of four action sequences. We reserved one sequence as the test set and used the remaining three for training. This setting aims to assess the model's ability to generalize to new actions from known subjects.

% In the more challenging unseen-subject evaluation setting, we constructed the validation set using the two subjects with the smallest amount of available data. From the remaining ten subjects, two were held out as the test set, while data from the other eight subjects were used for training. This setup is designed to assess the model’s generalization ability to completely unseen individuals.

To evaluate generalisation, we performed cross-validation on the model under two data splitting protocols: unseen-group and unseen-subject. In the unseen-group setting, for each subject (who performed four sequences), we hold out one sequence for testing and use the other three for training, evaluating generalization to new actions of known subjects. In the more challenging unseen-subject setting, we use the two subjects with the least data for validation, two other subjects for testing, and the remaining eight for training, testing generalization to completely new individuals.

\subsubsection{Evaluation Metrics.}
We evaluate the performance of human Mesh Reconstruction methods using four metrics: Procrustes-aligned mean per joint position error (PA-MPJPE), mean per joint position error (MPJPE), mean per vertex position  error (MPVE), and the 3D acceleration of the predicted 3D joints (Accel).

\subsection{Results}

% \vspace{-15pt}
\begin{table}[htbp]
  \caption{Overall Performance of MDP-Net on the unseen-group and unseen-subject dataset settings. MPJPE, PA-MPJPE and MPVE are in $cm$; Accel is in $mm/s^2$.}
  \label{performance}
  \centering
  \begin{tabular}{c|c|cccc}
    \Xhline{3\arrayrulewidth}
    Dataset Mode & Fold & MPJPE$\downarrow$ & PA-MPJPE$\downarrow$ & MPVE$\downarrow$ & Accel$\downarrow$ \\
    \Xhline{3\arrayrulewidth}
    \multirow{5}{*}{Unseen Group} & 1 & 11.8 & 8.3 & 13.6 & 17.4 \\
    \cline{2-6}
    & 2 & 12.8 & 8.4 & 14.8 & 18.5 \\
    \cline{2-6}
    & 3 & 13.4 & 8.2 & 15.0 & 17.8 \\
    \cline{2-6}
    & 4 & 12.4 & 8.0 & 14.6 & 18.1 \\
    \cline{2-6}
    & \textbf{Average} & \textbf{12.6} & \textbf{8.2} & \textbf{14.5} & \textbf{18.0} \\
    \hline
    \multirow{6}{*}{Unseen Subject} & 1 & 14.3 & 8.7 & 16.3 & 17.6 \\
    \cline{2-6}
    & 2 & 17.0 & 10.1 & 19.4 & 16.6 \\
    \cline{2-6}
    & 3 & 16.2 & 10.0 & 17.9 & 14.8 \\
    \cline{2-6}
    & 4 & 12.2 & 8.2 & 14.1 & 17.8 \\
    \cline{2-6}
    & 5 & 16.3 & 8.6 & 17.5 & 18.7 \\
    \cline{2-6}
    & \textbf{Average} & \textbf{15.2} & \textbf{9.1} & \textbf{17.0} & \textbf{17.1} \\
    \Xhline{3\arrayrulewidth}
  \end{tabular}
\end{table}
% \vspace{-10pt}

% \vspace{-15pt}
\begin{figure}[htbp]
\centerline{\includegraphics[width=1.0\linewidth]{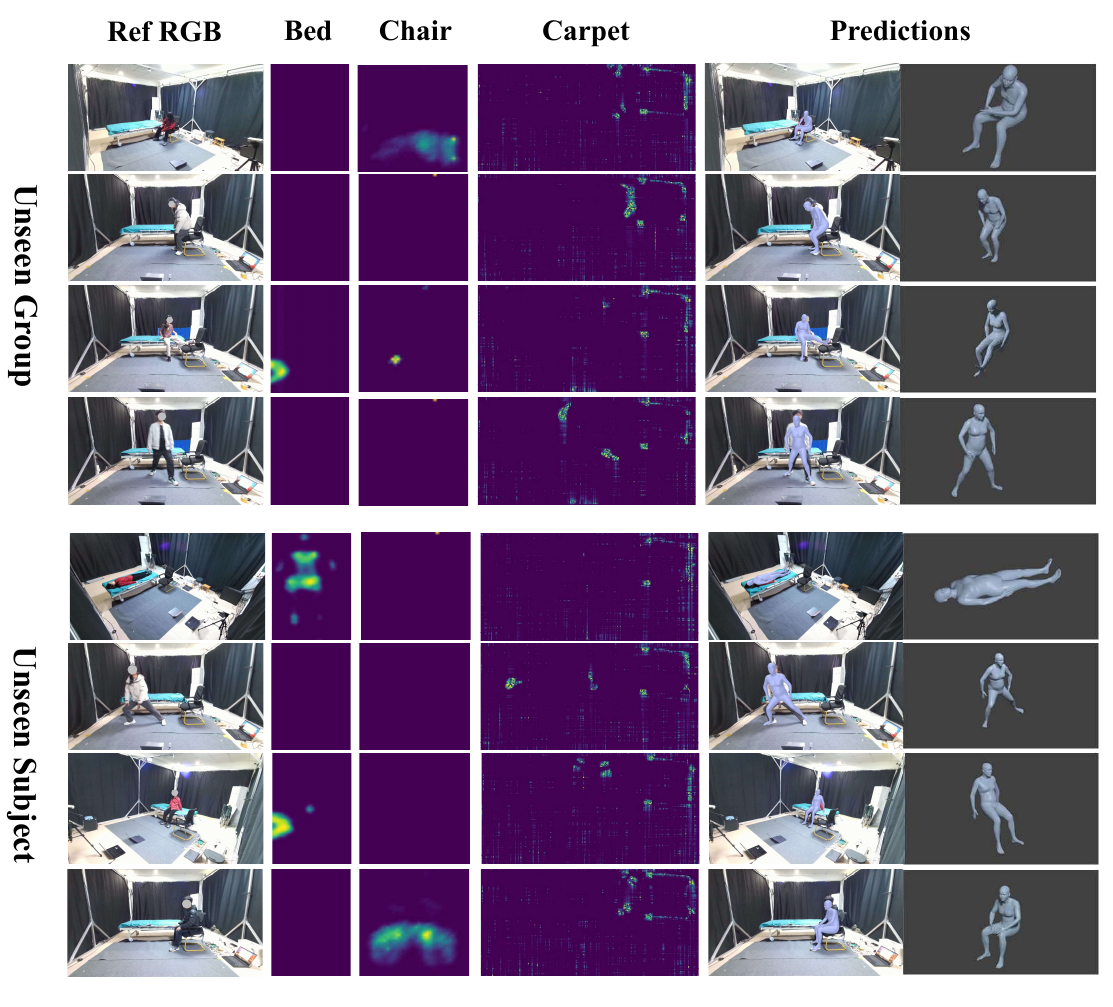}}
\caption{Qualitative results of MDP-Net on unseen group and unseen subject 
dataset. For each prediction, the left side displays a rendering generated using pre-calibrated camera parameters, while the right side provides a detailed view of the corresponding SMPL model from a novel perspective.}
\label{img:group}
\end{figure}
% \vspace{-20pt}

Table~\ref{performance} presents the quantitative evaluation results of our proposed MDP-Net on the 3D human mesh reconstruction task. Under the unseen-pose configuration, MDP-Net achieves optimal performance with an MPJPE of 12.6 $cm$, MPVE of 14.5 $cm$, and Accel of 18.0 $mm/s^2$, demonstrating robust reconstruction capability for unseen poses. In the more challenging unseen-subject setting, the model attains an MPJPE of 15.2 $cm$—only a marginal increase of 2.9 $cm$ compared to the unseen-pose setting—indicating strong adaptability and generalization ability across subjects with varying body shapes.

We also conducted a qualitative evaluation of the model. As shown in Fig~\ref{img:group}, our method consistently produces accurate and plausible human meshes across a variety of complex scenarios, including lying, sitting, and standing postures, dynamic transitions between poses, and multi-device interaction scenarios.

It is noteworthy that our method accurately reconstructs diverse human poses even for unseen subjects, as demonstrated in Fig~\ref{img:group} .

\subsubsection{Comparison with the single-device approach.}

\vspace{-15pt}
\begin{table*}[htbp]
\centering
\caption{Comparison of MDP-Net and single-device methods across different pose subsets. MPJPE and MPVE are in cm; Accel is in mm/s²}
\label{table1}
% 定义居中的X列类型
\renewcommand{\tabularxcolumn}[1]{m{#1}} % 垂直居中
\newcolumntype{C}{>{\centering\arraybackslash}X} % 水平居中
\begin{tabularx}{\textwidth}{c|C|C|C|C|C|C|C|C|C}
\Xhline{3\arrayrulewidth}
 \multirow{2}{*}{} & \multicolumn{3}{|c}{Lying Pose Subset} & \multicolumn{3}{|c}{Sitting Pose Subset} & \multicolumn{3}{|c}{Standing Pose Subset}\\
\cline{2-10}
& MPJPE & MPVE & Accel & MPJPE & MPVE & Accel & MPJPE & MPVE & Accel \\
\Xhline{3\arrayrulewidth}
Only Bed & 13.0 & 14.6 & \textbf{9.9} & - & - & - & - & - & - \\
\hline
Only Chair & - & - & - & 13.0 & - & 18.3 & - & - & - \\
\hline
Only Carpet & - & - & - & - & - & - & 19.6 & 22.9 & \textbf{12.2} \\
\hline
Ours & \textbf{8.5} & \textbf{9.9} & 13.8 & \textbf{8.6} & \textbf{9.8} & \textbf{16.3} & \textbf{14.5} & \textbf{16.4} & 18.8 \\
\Xhline{3\arrayrulewidth}
\end{tabularx}
\label{tab:single}
\end{table*}
\vspace{-5pt}

We divided the dataset into three subsets based on the primary pressure distribution device: the Lying Pose Subset, the Standing Pose Subset, and the Sitting Pose Subset. Each subset includes interaction poses where pressure data from multiple devices coexist. To establish single-device baselines, we selected representative comparative methods for each device as follows:
\begin{itemize}
\item[$\bullet$]\textbf{Bed:} We chose PiMesh~\cite{pimesh}, which has advanced performance in this field.
\item[$\bullet$]\textbf{Carpet:} Among existing methods, Luo et al.~\cite{Intelligentcarpet} output only 3D joints, not SMPL meshes; Cavatar~\cite{Cavatar} has not released code; MotionPro~\cite{motionpro} uses additional visual modalities, deviating from our pure‑pressure setting. For a fair SMPL-based comparison, and because carpet and mattress pressure share similar 2D image-like morphology, we adopt the PiMesh architecture—originally designed for mattress pressure—and retrain it on carpet data as our baseline.
% Among existing methods, Luo et al.~\cite{Intelligentcarpet} only predicts 3D joint positions, which is inconsistent with the SMPL mesh output format; Cavatar~\cite{Cavatar} has not open-sourced its code; and MotionPro~\cite{motionpro} incorporates visual modalities, which contradicts our pure pressure-input setting. To ensure a fair comparison under the unified task of SMPL estimation, and considering the similarity in signal morphology between carpet pressure and mattress pressure (both being 2D pressure distribution images), we adopted the same PiMesh architecture used for the mattress pressure baseline and retrained it on carpet data to establish a competitive baseline.
\item[$\bullet$]\textbf{Chair:} For chair pressure data, the method proposed by Zhao et al. ~\cite{zmj} is adopted as the baseline. Although this method outputs 3D joint positions rather than SMPL parameters, it is currently the only approach specifically designed for chair pressure input. Therefore, in seat posture estimation, the comparison is primarily conducted based on joint positions.
\end{itemize}

Quantitative results are shown in Table~\ref{tab:single}, where our method outperforms all single-device baselines across the three pose subsets. This outcome preliminarily indicates that multi-device fusion contributes to improved estimation accuracy.

\begin{figure}[htbp]
\centerline{\includegraphics[width=1.0\linewidth]{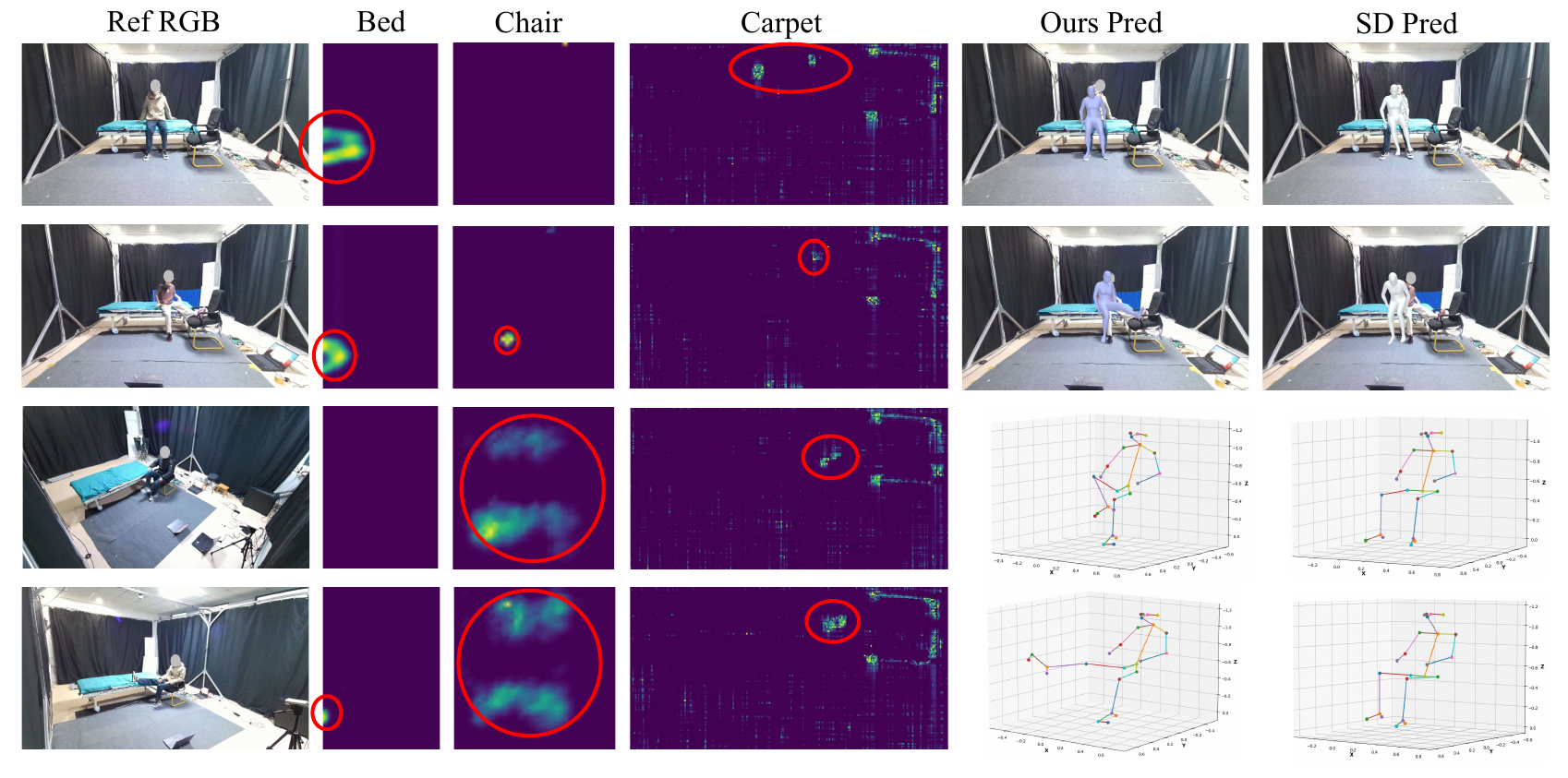}}
\caption{Comparison between Our Method and Single-Device Baseline Predictions. We highlight the pressure-activated regions by red ellipses. Pred: Predictions, SD: Single-Device Methods.}
\label{img:compare}
\end{figure}

Further qualitative analysis (Fig~\ref{img:compare}) reveals that in scenarios involving pressure from multiple devices, single-device methods—being limited to partial contact information—may yield unrealistic poses or suffer from global structural errors, such as mapping a local pressure distribution to an incorrect full-body pose. In contrast, our proposed multi-device fusion approach leverages global contact information comprehensively, demonstrating more stable estimation performance in multi-device scenarios. More analysis refers to the Sup. Mat..

\vspace{-15pt}
\vspace{-10pt}

\subsubsection{Ablation Study.}

Our ablation study systematically evaluates the impact of each core component in the model on performance, with quantitative results presented in Table~\ref{tab:ablation}. The experiment primarily analyzes three key design choices: the selection of the temporal encoder, the incorporation of the MoE architecture and the use of the regressor.

The results demonstrate that the Transformer-based temporal encoder significantly outperforms the GRU variant. The most notable performance drop occurs when the MoE architecture is removed and concatenated features are fed directly into the regressor, underscoring the critical role of MoE in feature fusion. Finally, we demonstrated the effectiveness of introducing a regressor, which enables the decoupling of responsibilities from the MoE. Experiments show that after removing the regressor, all performance metrics of the model decreased.

% During the model design phase, we experimented with predicting SMPL parameters directly from the MoE output; however, inspired by Cliff~\cite{cliff}, we ultimately introduced a regressor composed of MLP and Dropout layers as a decoder to extract SMPL parameters from the MoE-fused features. The experimental results validate the effectiveness of this regressor.

% We conducted an ablation study on the total number of experts and the number of activated experts, with results shown in Table~\ref{tab:expert}. Based on the  characteristics of our research task, we tested configurations with 3 and 5 experts: the 3-expert configuration was chosen based on the correspondence to the three device types, while the 5-expert configuration was designed to cover five fundamental motion patterns—standing, sitting, lying, standing-to-sitting transition, and lying-to-sitting transition. Experimental results indicate that the model achieves optimal performance when the total number of experts is 5, with either 3 or 5 experts activated. Considering practical applicability and efficiency, we ultimately selected the configuration with 3 activated experts. This setup maintains performance while effectively controlling the model's parameter size and improving inference speed.

We conducted an ablation study on the total number of experts and the number of activated experts, with results shown in Table~\ref{tab:expert}.  Aligned with our task, we test two settings: 3 experts (matching the three device types) and 5 experts (covering five basic motion patterns: standing, sitting, lying, stand-to-sit, and lie-to-sit). Results show that the model performs best with 5 total experts and 3 or 5 activated. Considering practical efficiency, we adopt the setting with 3 activated experts, which maintains performance while controlling parameters and improving inference speed.

\vspace{-10pt}
\begin{table}[htbp]
  \caption{Quantitative results of different model variants.}
  \label{tab:ablation}
  \centering
  \begin{tabular}{c|cccc}
    \Xhline{3\arrayrulewidth}
     & MPJPE$\downarrow$ & PA-MPJPE$\downarrow$ & MPVE$\downarrow$ & Accel$\downarrow$\\
    \Xhline{3\arrayrulewidth}
    Base & \textbf{11.8} & \textbf{8.3} & \textbf{13.6} & \textbf{17.4} \\
    \hline
    GRU & 13.3 & 8.4 & 15.4 & 32.6 \\
    \hline
    w/o MoE & 14.8 & 9.4 & 16.6 & 23.4 \\
    \hline
    % w/o 双重循环 & 11.8 & 8.1 & 13.7 & 18.4 \\
    % \hline
    w/o Regressor & 12.5 & 8.4 & 14.1 & 19.9 \\
    \Xhline{3\arrayrulewidth}
  \end{tabular}
\end{table}
\vspace{-10pt}

\vspace{-10pt}
\begin{table}[]
  \caption{Performance Comparison of MDP-Net with Different Expert Configurations. N: Total number of the experts, K: Number of activated experts}
  \label{tab:expert}
  \centering
  \begin{tabular}{c|ccc|ccccc}
    \Xhline{3\arrayrulewidth}
    \multirow{2}{*}{Configuration} & \multicolumn{3}{c}{$N=3$} & \multicolumn{5}{|c}{$N=5$}\\
    \cline{2-9}
     & $K=1$ & $K=2$ & $K=3$ & $K=1$ & $K=2$ & $K=3$ & $K=4$ & $K=5$ \\
    \hline
    MPJPE & 12.8 & 12.3 & 12.2 & 12.0 & 12.1 & \textbf{11.8} & 12.3 & \textbf{11.8} \\
    \Xhline{3\arrayrulewidth}
  \end{tabular}
\end{table}
\vspace{-10pt}

\section{Conclusion}
% In this paper, we proposed an end-to-end neural network, MDP-Net, capable of directly regressing human mesh parameters from multi-device temporal pressure sequences. To address the inherent heterogeneity and complementarity in multi-source pressure information, the proposed method incorporates a Mixture-of-Experts (MoE) module for dynamic and adaptive fusion of multi-device features, thereby offering a novel approach for vision-free, privacy-preserving human activity sensing. To address the data scarcity challenge for this task, we synchronously collected a dataset comprising multi-device pressure sequences and multi-view RGB images, covering 12 subjects and containing 672K RGB images and 96K synchronized multi-device pressure data samples. High-quality 2D/3D joint annotations and SMPL mesh labels were generated via a multi-view optimization process. Experimental results on this dataset demonstrate that MDP-Net achieves a mean per-joint position error (MPJPE) of 12.6 cm. This work not only validates the effectiveness of multi-device pressure information for human pose and shape estimation, but also provides a feasible technical pathway for applications in privacy-sensitive scenarios such as rehabilitation monitoring and in-home sensing.

This paper presents MDP‑Net, an end‑to‑end network that regresses 3D human mesh parameters directly from multi‑device temporal pressure sequences. To handle the inherent heterogeneity and complementarity of multi‑source pressure data, we incorporate a Mixture‑of‑Experts (MoE) module for dynamic, adaptive feature fusion, offering a new vision‑free and privacy‑preserving sensing approach. To overcome data scarcity, we collect a synchronized dataset of multi‑view RGB images and multi‑device pressure sequences from 12 subjects, comprising 672K images and 96K pressure samples, with high‑quality 2D/3D joint and SMPL mesh labels generated via multi‑view optimization. Experiments show MDP‑Net achieves 12.6$cm$ MPJPE, demonstrating the effectiveness of multi‑device pressure fusion for pose and shape estimation, and providing a viable technical path for privacy‑sensitive applications like rehabilitation and in‑home monitoring.

%
% ---- Bibliography ----
%
% BibTeX users should specify bibliography style 'splncs04'.
% References will then be sorted and formatted in the correct style.
%
%

\end{document}